\documentclass{article} 
\usepackage[latin1]{inputenc}
\usepackage{nips13submit_e,times}

\usepackage[dvips]{graphicx}
\usepackage{color}
\usepackage{mathrsfs}

\usepackage{textcomp,amssymb,amsmath,bm}
\usepackage[
  backend=bibtex,
  firstinits=true,
  dashed=false,
  style=authoryear,
  isbn=false,
  maxcitenames=2,
  maxbibnames=100,
]{biblatex}
\DeclareFieldFormat{sentencecase}{\MakeSentenceCase*{#1}}
\renewbibmacro{title}{%
    \ifthenelse{\iffieldundef{title}\AND\iffieldundef{subtitle}}{}
        {\ifthenelse{\ifentrytype{article}\OR\ifentrytype{inbook}%
            \OR\ifentrytype{incollection}\OR\ifentrytype{inproceedings}}
            {\printtext[title]{%
                \printfield[sentencecase]{title}%
                \setunit{\subtitlepunct}%
                \printfield[sentencecase]{subtitle}}%
                \newunit}%
            {\printtext[title]{%
                \printfield[titlecase]{title}%
                \setunit{\subtitlepunct}%
                \printfield[titlecase]{subtitle}}%
                \newunit}}%
    \printfield{titleaddon}}

\usepackage{hyperref}
\usepackage{url}
\bibliography{ladder_all}

\newcommand{\tr}[0]{{\mathrm{tr}}}
\newcommand{\sigmoid}[0]{{\mathrm{sigmoid}}}
\newcommand{\vect}[1]{{\mathbf{#1}}}
\newcommand{\matr}[1]{{\mathbf{#1}}}
\newcommand{\set}[1]{{\mathbb{#1}}}
\newcommand{\HLL}[0]{{\hspace{-2mm}\begin{minipage}{3cm}
\centering
\bf
Hierarchical\\
latent variable\\
model
\end{minipage}}}
\newcommand{\SAE}[0]{{\hspace{-2mm}\begin{minipage}{3cm}
\centering
\bf
Standard\\
autoencoder\\
network
\end{minipage}}}
\newcommand{\Ladder}[0]{{\hspace{-2mm}\begin{minipage}{3cm}
\centering
\bf
Ladder\\
autoencoder\\
network
\end{minipage}}}
\newcommand{\CorruptPath}[0]{{\hspace{-2mm}\begin{minipage}{3cm}
\centering
\bf
Corrupted \\
$f$ path
\end{minipage}}}
\newcommand{\DenoisingPath}[0]{{\hspace{-2mm}\begin{minipage}{3cm}
\centering
\bf
Denoising \\
$g$ path
\end{minipage}}}
\newcommand{\CleanPath}[0]{{\hspace{-2mm}\begin{minipage}{3cm}
\centering
\bf
Clean \\
$f$ path
\end{minipage}}}

\title{From Neural PCA to Deep Unsupervised Learning}

\author{Harri Valpola \\
ZenRobotics Ltd.\\
Vilhonkatu 5 A\\
00100 Helsinki, Finland \\
\texttt{harri@zenrobotics.com}}

\nipsfinalcopy 

\begin{document}

\maketitle

\begin{abstract}
A network supporting deep unsupervised learning is presented.  The
network is an autoencoder with lateral shortcut
connections from the encoder to decoder at each level of the hierarchy.
The lateral shortcut connections allow the higher
levels of the hierarchy to focus on abstract invariant features.
While standard autoencoders are analogous to latent variable models with a
single layer of stochastic variables, the proposed network is analogous
to hierarchical latent variables models.

Learning combines denoising autoencoder and denoising sources
separation frameworks.
Each layer of the network contributes to the cost function a term
which measures the distance of the representations produced by the
encoder and the decoder.
Since training signals originate from all levels of the network, all
layers can learn efficiently even in deep networks.

The speedup offered by cost terms from higher levels of the hierarchy
and the ability to learn invariant features are demonstrated in
experiments.
\end{abstract}


\section{Introduction}

Ever since \textcite{Hubel:62} published their findings about a
hierarchy of increasingly abstract invariant visual features in the
cat neocortex, researchers have been
trying to mimic the same hierarchy of feature extraction stages with
artificial neural networks.
An early example is the neocognitron by
\textcite{Fukushima:1979neocognitron}.

Nowadays the art of estimating deep feature extraction hierachies is
called deep learning \parencite[for a thorough overview of the history
  of the field, including recent developments,
see the review by][]{Schmidhuber2015}. The topic has received a
lot of attention after \textcite{HinSal06,hinton:06afast} proposed an
unsupervised\footnote{Unsupervised learning aims at representing
  structure in the input data, often by means of features.  The
  resulting features can be used as input for classification tasks or
  as initialization for further supervised learning.}  pre-training
scheme which made subsequent supervised learning efficient for a
deeper network than before.

What is somewhat embarrasing for the field, though, is that recently
purely supervised learning has achieved as good or better results as
unsupervised pre-training schemes
\parencite[e.g.,][]{ciresan:2010,Krizhevsky2012NIPS}.  In most
classification problems, finding and producing labels for the samples
is hard. In many cases plenty of unlabeled data exist and it seems
obvious that using them should improve the results. For instance,
there are plenty of unlabeled images available and in most image classification
tasks there are vastly more bits of information in the statistical
structure of input images than in their
labels\footnote{Consider, for
  example, the ImageNet classification problem that
  \textcite{Krizhevsky2012NIPS} tackled. With 1000 target classes, each
  label carries less than 10 bits of information. Compare this with
  the amount of information contained in the $256 \times 256$ RGB
  images used as input. It is impossible to say exactly how many bits of
  information each image carries but certainly several
  orders of magnitude more than 10 bits.}.

It is argued here that the reason why unsupervised learning has not been able to
improve results is that most current versions are incompatible with
supervised learning. The problem is that many unsupervised learning
methods try to represent as much information about the original
data as possible whereas supervised learning tries to filter out all
the information which is irrelevant for the task at hand.

This chapter presents an unsupervised learning network whose
properties make it a good fit with supervised learning. First,
learning is based on minimizing a cost function much the same way
as in stochastic gradient descent of supervised feedforward networks.
Learning can therefore continue alongside supervised learning rather
than be restricted to a pre-training phase.
Second, the network can discard information from the higher layers and
leave the details for lower layers to represent. This means that the approach
allows supervised learning to select the relevant features. The proposed
unsupervised learning can filter out noise from the selected features and
come up with new features that are related to the selected features.

The two main new ideas presented in this chapter are as follows:
\begin{enumerate}
\item Section~\ref{sec:val-ladder} explains how adding lateral
  shortcut connections to an autoencoder gives it
  the same representational capacity as hierarchical latent variable
  models. This means that higher layers no longer need to represent
  all the details but can concentrate on abstract invariant
  representations. The model structure is called a ladder network because
  two vertical paths are connected by horizontal lateral connections at regular
  intervals.
\item Learning of deep autoencoders can be slow since training signals
  need to travel a long distance from the decoder output through both
  the decoder and encoder. Lateral shortcuts tend to slow it down even
  further since the shortcuts learn first, shunting the training
  signals along the longer paths. Section~\ref{sec:val-denoising}
  explains how this can be remedied by adding training targets to each
  level of the hierarchy. The novel idea is to combine denoising
  source separation (DSS) framework \parencite{Sarela:2005} with
  training denoising functions to remove injected noise
  \parencite{Vincent2008}.
\end{enumerate}

The experiments presented in Section~\ref{sec:val-experiments} demonstrate
that the higher levels of a ladder network can discard information and
focus on invariant representations and that the training targets on
higher layers speed up learning.  The results presented here are
promising but preliminary.  Section~\ref{sec:val-discussion} discusses
potential extensions and related work.

\section{Ladder network: an autoencoder which can discard information}
\label{sec:val-ladder}

As argued earlier, unsupervised learning needs to tolerate discarding
information in order to work well with supervised learning.  Many
unsupervised learning methods are not good at this but one class of
models stands out as an exception: hierarchical latent variable
models.  Unfortunately their derivation can be quite complicated and
often involves approximations which compromise their performance.  A
simpler alternative is offered by autoencoders which also have the
benefit of being compatible with standard supervised feedforward
networks.  They would be a promising candidate for combining
supervised and unsupervised learning but unfortunately autoencoders
normally correspond to latent variable models with a single layer of
stochastic variables, that is, they do not tolerate discarding
information.

This section summarizes the complementary roles of supervised and
unsupervised learning, reviews latent variable models and their
relation to standard autoencoder networks and proposes a new network
structure, the ladder network, whose lateral shortcut connections give
it the same representational capacity as hierarchical latent variable
models.

\subsection{Complementary roles of supervised and unsupervised learning}

Consider the roles of supervised and unsupervised learning in a
particular task, such as classification, prediction or regression.
Further assume that 1) there are input-output pairs which can be used
for supervised learning but far more unlabeled samples that are
lacking the output, with only the inputs available and 2) inputs have
far more information than the outputs.

In general, this typical setup means that unsupervised learning should
be used for anything it works for because the precious bits of
information in the available output samples should be reserved for those
tasks that unsupervised learning cannot handle.

The main role for supervised learning is clear enough: figure out which
type of representations are relevant for the task at hand.  Only
supervised learning can do this because, by definition, unsupervised
learning does not have detailed information about the task.

One obvious role, the traditional one, for unsupervised learning is to
act as a pre-processing or pre-training step for supervised learning.
However, the key question is: what can unsupervised learning do after
supervised learning has kicked in. This is important because in many
problems there is so much information in the inputs that it cannot
possibly be fully summarized by unsupervised learning first. Rather,
it would be useful for unsupervised learning to continue tuning the
representations even after supervised learning has started to tune the
relevant features and filter out the irrelevant ones.

The combination of supervised and unsupervised learning is known
as semi-supervised learning.  As argued earlier, it can be a happy
marriage only if unsupervised learning is content with discarding
information and concentrating on the features which supervised learning
deems relevant.

What unsupervised learning should be able to do efficiently is to find
new features which correlate with and predict the features selected by
supervised learning. This improves generalization to new samples. As
an example, consider learning to recognize a face. Suppose supervised
learning has figured out that an eye is an important feature for
classifying faces vs. non-faces from a few samples. What unsupervised
learning can do with all the available unlabeled samples, is find
other features which correlate with the selected one. Such new
features could be, for instance, a detector for nose, eye brow, ear,
mouth, and so on. These features improve the generalization of a face detector
in cases where the eye feature is missing, for instance due to eyes being
closed or occluded by sunglasses.

Specifically, what unsupervised learning must \textit{not} do is keep pushing
new features to the representation intended for supervised learning
simply because these features carry information about the inputs.
While such behavior may be reasonable as a pre-training or
pre-processing step, it is not compatible with semi-supervised
learning. In other words, before unsupervised learning knows which
features are relevant, it is reasonable to select features which carry
as much information as possible about the inputs. However, after
supervised learning starts showing a preference of some features over
some others, unsupervised learning should follow suite and present
more of the kind that supervised learning seems to be interested in.

\subsection{Latent variable models}

Many unsupervised learning methods can be framed as latent variable
models \parencite[see, e.g.,][]{Bishop99LVM}
where unknown latent varibles $\vect{s}(t)$ are assumed to
generate the observed data $\vect{x}(t)$.  A common special case with
continuous variables is that the latent variables predict the mean
of the observations:
\begin{equation}
  \label{eq:val-1}
  \vect{x}(t) = g(\vect{s}(t); \vect{\xi}) + \vect{n}(t) \, ,
\end{equation}
where $\vect{n}(t)$ denotes the noise or modeling error and
$\vect{\xi}$ the parameters of mapping $g$.  Alternatively, the same can be
expressed through a probability model
\begin{equation}
  \label{eq:val-2}
  p_{\mathrm{x}}(\vect{x}(t) | \vect{s}(t), \vect{\xi}) = 
    p_{\mathrm{n}}(\vect{x}(t) - g(\vect{s}(t); \vect{\xi})) \, ,
\end{equation}
where $p_{\mathrm{n}}$ denotes the probability density function of the noise term
$\vect{n}(t)$.  Inference of the unknown latent variables $\vect{s}(t)$
and parameters $\vect{\xi}$ can then be based simply on minimizing the
mismatch between the observed $\vect{x}(t)$ and its reconstruction
$g(\vect{s}(t); \vect{\xi})$, or more generally on probabilistic modeling.

The models defined by Eqs.~\eqref{eq:val-1} or \eqref{eq:val-2} have just
one layer of latent variables which tries to represent everything
there is to represent about the data. Such models have trouble letting
go of any piece of information since this would increase the
reconstruction error.  Also, in many cases an abstract invariant
feature (such as ``a face'') cannot reduce the reconstruction error
alone without plenty of accompanying details such as position,
orientation, size, and so on.  All of those details need to be represented
alongside the relevant feature to show any benefit in reducing the
reconstruction error.  This means that latent variable models with a
single layer of latent variables have trouble discarding information
and focusing on abstract invariant features.

It is possible to fix the situation by introducing a hierarchy of
latent variables:
\begin{equation}
  \label{eq:val-3}
  p(\vect{s}^{(l)}(t) | \vect{s}^{(l+1)}(t), \vect{\xi}^{(l)}) \, ,
\end{equation}
where the superscript $(l)$ refers to variables on layer $l$. The observations
can be taken into the equation by defining $\vect{s}^{(0)} := \vect{x}$.
Now the latent variables on higher levels no longer need to represent
everything.  Lower levels can take care of representing details while
higher levels can focus on selected features, abstract or not.

In such hierarchical models, higher-level latent variables can still
represent just the mean of the lower-level variables,
\begin{equation}
  \label{eq:val-3b}
  \vect{s}^{(l)}(t) = g^{(l)}(\vect{s}^{(l+1)}(t); \vect{\xi}^{(l)})
    + \vect{n}^{(l)}(t) \, ,
\end{equation}
but more generally, the higher-level variables can represent any
properties of the distribution, such as the variance
\parencite{Valpola04SigProc}.  For binary variables,
sigmoid units are often used for representing the dependency
\parencite[for a recent example of such a model,
  see][]{Gregor2014ICML}.

Exact inference, that is, computing the posterior probability of the
unknown variables (latent variables and the parametes of the mappings)
is typically mathematically intractable. Instead, approximate
inference techniques such as variational Bayesian methods are employed
\parencite[e.g.,][]{Valpola04SigProc,Gregor2014ICML}.  They amount to
approximating the intractable exact posterior probability with a
simpler tractable approximation.  Learning then corresponds to
iteratively minimizing the cost function with respect to the posterior
approximation.  For instance, in the case of continuous latent
variables, the posterior could be approximated as Gaussian with a
diagonal covariance.  For each unknown variable, the mean and variance
would then be estimated in the course learning.  These posterior means
and variances typically depend on the values on both lower and higher
layers and inference would therefore proceed iteratively.

If hierarchical latent variable models were easy to define and learn,
the problem would be solved. Unfortunately hierarchical models often
require complex probabilistic methods to train them.
They often involve approximations which
compromise their performance~\parencite{Ilin03ICA} or are limited to
restricted model structures which are mathematically tractable. Also, many
training schemes require the latent variable values to be updated
layer-wise by combining bottom-up information with top-down
priors. This slows down the propagation of information in the network.

\subsection{Autoencoders and deterministic and stochastic latent variables}

Autoencoder networks resemble in many ways single-layer latent
variable models.  The key idea is that the inference process of
mapping observations $\vect{x}(t)$ to the corresponding latent
variables, now called hidden unit activations $\vect{h}(t)$, is
modeled by an encoder network $f$ and the mapping back to observations
is modeled by a decoder network $g$:
\begin{eqnarray}
  \label{eq:val-4}
  \vect{h}(t) & = & f(\vect{x}(t); \vect{\xi}_f) \\
  \vect{\hat{x}}(t) & = & g(\vect{h}(t); \vect{\xi}_g) \, .
  \label{eq:val-5}
\end{eqnarray}
The mappings $f$ and $g$ are called encoder and decoder mappings,
respectively. In connection to latent variable models, analogous
mappings are called the recognition and reconstruction mappings.

Learning of autoencoders is based on minimizing the difference between
the observation vector $\vect{x}(t)$ and its reconstruction
$\vect{\hat{x}}(t)$, that is, minimizing the cost $||\vect{x}(t) -
\vect{\hat{x}}(t)||^2$ with respect to the parameteres $\vect{\xi}_f$
and $\vect{\xi}_g$.  For the remainder of this chapter, all mappings
$f$ and $g$ are assumed to have their own parameters but they are
omitted for brevity.

Just like latent variable models, autoencoders can be stacked together:
\begin{eqnarray}
  \label{eq:val-6}
  \vect{h}^{(l)}(t) & = & f^{(l)}(\vect{h}^{(l-1)}(t)) \\
  \vect{\hat{h}}^{(l-1)}(t) & = & g^{(l)}(\vect{\hat{h}}^{(l)}(t)) \, .
  \label{eq:val-7}
\end{eqnarray}
As before, the observations are taken into the equation by defining
$\vect{h}^{(0)} := \vect{x}$. Furthermore, now $\vect{\hat{h}}^{(L)}
:= \vect{h}^{(L)}$ for the last layer $L$, connecting the
encoder and decoder paths.

Typically over the course of learning, new layers are added to the
previously trained network. After adding and training the last layer,
training can continue in a supervised manner using just the mappings
$f^{(l)}$, which define a multi-layer feedforward network, and minimizing
the squared distance between the actual outputs $\vect{h}^{(L)}$ and
desired targets outputs.

It is tempting to assume that the hierarchical version of
the autoencoder in Eqs.~(\ref{eq:val-6}--\ref{eq:val-7}) corresponds somehow to
the hierarchical latent variable model in Eq.~\eqref{eq:val-3b}.
Unfortunately this is not the case because the intermediate hidden
layers $0 < l < L$ act as so called deterministic variables while the
hierarchical latent variable model requires so called stochastic
variables. The difference is that stochastic variables have
independent representational capacity. No matter what the priors
tell, stochastic latent variables $\vect{s}^{(l)}$ can overrule this
and add their own bits of information to the reconstruction.
By contrast, deterministic variables such as $\hat{h}^{(l)}$ add zero bits of
information and, assuming the deterministic mappings $g^{(l)}$ are
implemented as layered networks, correspond to the hidden layers of
the mappings $g^{(l)}$ between the stochastic variables
$\vect{s}^{(l)}$ in Eq.~\eqref{eq:val-3}.

In order to fix this, we are going to take cue from the inference
structure of the hierarchical latent variable model in
Eq.~\eqref{eq:val-3}.  The
main difference to Eq.~\eqref{eq:val-7} is that inference of
$\vect{\hat{s}}^{(l)}(t)$ combines information from bottom-up
likelihood and top-down prior but Eq.~\eqref{eq:val-7} only depends on
top-down information. In other words, $\vect{\hat{h}}(t)$ in
Eq.~\eqref{eq:val-7} cannot add any new information to the representation
because it does not receive that information from the bottom-up path.
The fix is to add a shortcut connection from the bottom-up encoder
path to the modified top-down decoder path:
\begin{equation}
  \vect{\hat{h}}^{(l-1)}(t) = g^{(l)}(\vect{\hat{h}}^{(l)}(t), \vect{h}^{(l-1)}(t)) \, .
  \label{eq:val-shortcut}
\end{equation}
Now $\vect{\hat{h}}^{(l)}$ can recover information which is missing in
$\vect{\hat{h}}^{(>l)}$. In other words, the higher layers do not need
to represent all the details. Also, the mapping $g^{(l)}$ can learn to
combine abstract information from higher levels, such as ``face'',
with detailed information about position, orientation, size, and so
on, from lower layers\footnote{The shortcut connections are only part
  of the solution since adding them actually short-circuits standard
  autoencoder learning: mapping $g^{(l)}$ can simply copy the
  shortcut input $\vect{h}^{(l-1)}(t)$ into its output so there is no
  incentive to use any of the higher layers. This is related to the
  inability of standard autoencoder training to learn over-complete
  representations. Fortunately, denoising autoencoders can overcome
  this problem simply by adding noise to inputs as will be explained
  in Section~\ref{sec:val-denoising-autoencoders}.}.
This means that the higher layers can focus on
representing abstract invariant features if they seem more relevant
to the task at hand than the more detailed information.

\begin{figure}
  \centering
  \hspace*{-8mm}\input{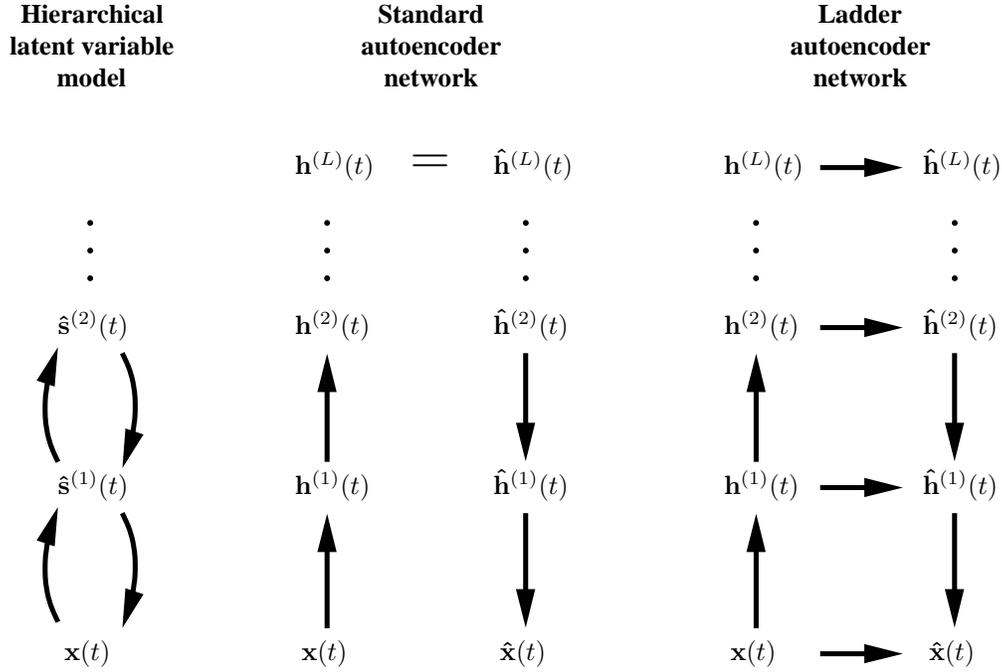} 
  \caption{The inference structure of a hierarchical latent variable model is
    compared with the standard autoencoder and the proposed ladder
    network.  The details of inference in latent variable models are
    often complex and the posterior distribution approximation is more
    complex than just the posterior mean $\vect{\hat{s}}^{(l)}(t)$, but
    overall the picture is approximately as shown in the left.
    Since all information in the standard autoencoder
    network has to go through the highest layer, it needs to represent
    all the details of the input $\vect{x}(t)$.  Intermediate hidden
    layer activations $\vect{\hat{h}}^{(l)}(t)$ cannot independently
    represent information because they only receive information from
    the highest layer. In the ladder network, by constrast, lateral
    connections at each layer give a chance for each
    $\vect{\hat{h}}^{(l)}(t)$ to represent information independently from
    the higher layers. Also, abstract invariant representations at the
    higher levels can be interpreted in the context of detailed information
    without the higher levels having to represent all the details.}
  \label{fig:val-latent-auto}
\end{figure}

Figure~\ref{fig:val-latent-auto} shows roughly the inference structure of a
hierarchical latent variable model and compares it with a standard
autoencoder and the ladder network.  Note that while
$\vect{\hat{h}}^{(l)}(t)$ combines information both from bottom-up and
top-down paths in the ladder network,
$\vect{h}^{(l)}(t)$ does not. This direct path from
inputs to the highest layer means that training signals from the
highest layers can propagate directly through the network in the same
way as in supervised learning.  Gradient propagation already combines
information from bottom-up activations and top-down gradients so there
is no need for extra mixing of information.

\section{Parallel learning on every layer}
\label{sec:val-denoising}

A general problem with deep models which have an error function only
at the input layer (autoencoder) or at the output layer (supervised
feedforward models) is that many parts of the network are far away
from the source of training signals.  In fact, if the ladder model
shown in Fig.~\ref{fig:val-latent-auto} is trained in the same
fashion as regular autoencoders, that is, by minimizing the difference
between $\vect{x}(t)$ and $\vect{\hat{x}}(t)$, the problem only
becomes worse. That is because each shortcut connection has a chance
of contributing to the reconstruction $\vect{\hat{x}}(t)$, leaving a
shrinking share of the error for the higher layers.  Standard
autoencoders force all training signals to pass through all levels in
the hierarchy and even then learning through multiple layers of
nonlinear functions is difficult and slow.

By contrast, hierarchical latent variable models have cost functions for all
stochastic variables. Since the ladder network shares many
properties with hierarchical latent variable models, it seems
reasonable to try to find a way to introduce training signals at
each level of the hierarchy of the ladder network.
This section shows how this can be done
by combining denoising source separation (DSS) framework
\parencite{Sarela:2005} with training denoising functions to remove
injected noise \parencite{Vincent2008}.

\subsection{From neural PCA to denoising source separation}

In order to develop a system where learning is distributed
rather than guided by gradients propagating from a single error term, we
shall turn our attention to competitive unsupervised learning.

The starting point for the algorithms we will study is the neural principal
component analysis (PCA) learning rule by \textcite{Oja:1982} which can
utilize second order statistics of the input data to find principal
component projections.  When slight nonlinear modifications are made
to the learning rule and input data are whitened, the method becomes
sensitive to higher-order statistics and performs independent
component analysis (ICA) \parencite{Oja97NC}.

The nonlinearity used
in the algorithm can be interpreted as a constrast function which
measures the non-Gaussianity of the source distribution.  This is
the interpretation originally given to the popular FastICA algorithm
\parencite{Hyvarinen97NC}.  However, there is an alternative view:
the nonlinearity can be interpreted as a denoising function.
\textcite{Hyvarinen98NC} derived this as a maximum likelihood
estimate but we are going to follow the derivation by
\textcite{ValpPaju00ICA} who showed that the nonlinearity can be
interpreted as the expectation step of the expectation maximization algorithm.
Overall, nonlinear PCA learning
rule, combined with input whitening and orthogonalization of the projections,
can be interpreted as an efficient approximation to the expectation
maximization (EM) algorithm applied to a linear latent variable model tuned
for ICA \parencite{ValpPaju00ICA}.  This interpretation led to the
development of denoising source separation framework \parencite{Sarela:2005}.

The EM algorithm \parencite{Dempster77} is a method for optimizing the
parametric mappings of latent variable models. It operates by
alternating the E step (expectation of latent variables) and M step
(maximization of likelihood of the parameters).  The E step assumes
the mapping fixed and updates the posterior distribution of
$\vect{s}(t)$ for all $t$ while the M step does the reverse, updating
the mapping while assuming the posterior distribution of $\vect{s}(t)$
fixed.

The derivation by \textcite{ValpPaju00ICA} assumed a linear
reconstruction model
\begin{equation}
  \vect{\hat{x}}(t) = g^{(0)}(\vect{s}(t)) = \matr{A} \vect{s}(t) \, ,
  \label{eq:val-linear}
\end{equation}
which means that the E step boils down to
\begin{equation}
  \vect{\hat{s}}(t) = g^{(1)}(\matr{A}^{-1}\vect{x}(t)) =
  g^{(1)}(\vect{s}_0(t)) \, ,
  \label{eq:val-DSS-E}
\end{equation}
where we have denoted $\vect{s}_0(t) = \matr{A}^{-1}\vect{x}(t)$.
The mapping $g^{(1)}$ depends on the prior distribution of $\vect{s}(t)$ and
the noise distribution $p_{\mathrm{n}}$.  When the noise has low
variance $\sigma_{\mathrm{n}}^2$, it can be approximated as
\begin{equation}
  \vect{\hat{s}}(t) = g^{(1)}(\vect{s}_0(t)) \approx \vect{s}_0(t) +
  \sigma_{\mathrm{n}}^2 \frac{\partial \log p_\mathrm{s}(\vect{s}(t))}
        {\partial \vect{s}(t)}_{\vect{s}(t) = \vect{s}_0(t)} \, .
  \label{eq:val-DSS-denoise}
\end{equation}
The M step amounts to solving the regression problem in
Eq.~\eqref{eq:val-linear} with $\vect{\hat{s}}(t)$ substituting
$\vect{s}(t)$, that is, minimizing the cost
\begin{equation}
  C = \frac{1}{T} \sum_{t = 1}^T ||\vect{x}(t) -
    g^{(0)}(\vect{\hat{s}}(t))||^2 \, .
  \label{eq:val-C-EM}
\end{equation}
The problem with the above EM algorithm for estimating the model is
that its convergence scales with $\sigma_{\mathrm{n}}^2$.  Without noise,
the algorithm stalls completely.

\textcite{ValpPaju00ICA} showed that this can be remedied by considering
the fixed point of the algorithm.  At the same time, it is useful to
parametrize the inverse mapping
\begin{equation}
  \vect{s}_0(t) = f^{(1)}(\vect{x}(t)) = \matr{W} \vect{x}(t)
\end{equation}
and for M step minimize
\begin{equation}
  C = \frac{1}{T} \sum_{t = 1}^T ||\vect{s}_0(t) - \vect{\hat{s}}(t)||^2 =
  \frac{1}{T} \sum_{t = 1}^T ||f^{(1)}(\vect{x}(t)) - \vect{\hat{s}}(t)||^2
  \label{eq:val-C-DSS}
\end{equation}
instead of Eq.~\eqref{eq:val-C-EM}.  Equation~\eqref{eq:val-C-DSS}
needs to be accompanied by constraints on the covariance of
$\vect{s}_0(t)$ because the trivial solution $\matr{W} = \matr{0}$
and $\vect{\hat{s}}(t) = \vect{0}$ yields $C = 0$.

The algorithm resulting from these assumptions is essentially the same
as the nonlinear PCA learning rule.  When the input data are whitened,
the M step becomes simple matrix multiplication just as in the
nonlinear PCA learning rule, amounting to essentially Hebbian
learning.  The nonlinearity $g^{(1)}(\vect{s}_0(t))$ has the
interpretation that it is the expected value of the latent variables
given noisy observations, that is denoising.

What will be crucial for our topic, learning deep models, is that the
cost function~\eqref{eq:val-C-DSS} does not directly refer to the
input $\vect{x}(t)$ but only to the latent variable $\vect{s}(t)$ and its
denoised version.  In a hierarchical model this will mean that each
layer contributes terms to the cost function, bringing the source of
training signals close to the parameters on each layer.

Just as with the nonlinear PCA learning rule, there needs to be an additional
constraint which implements competition between the latent variables
because they could otherwise all converge to the same values.  In case of a
linear model, the easiest approach is to require $\matr{W}$ to be
orthogonal.  This does not apply to nonlinear models.  Instead, it is
possible to require that the covariance matrix of the latent variables
$\vect{s}(t)$ is a unit matrix \parencite{Almeida06}.

\subsection{Denoising autoencoders and generative stochastic networks}
\label{sec:val-denoising-autoencoders}

The denoising function used in Eq.~\eqref{eq:val-DSS-E} can be derived
from the prior distribution $p_{\mathrm{s}}$ of the latent variables.
There are many techniques for learning such distributions but a
particularly useful technique that directly learns the denoising
function was proposed by \textcite{Vincent2008} in connection with
autoencoders.  The idea is to corrupt the inputs fed into the
autoencoder with noise and ask the network to reconstruct the original
uncorrupted inputs.  This forces the autoencoder to learn how to
denoise the corrupted inputs.

\Textcite{Bengio2013gsn} further showed that it is possible to sample
from these so called denoising autoencoders
simply by iterating corruption and denoising.  The
distribution of the denoised samples converges to the original data
distribution because during training, the denoising function learns to
cancel the diffusion resulting from the corruption of input data.  The
diffusive forces are proportional to $-\frac{\partial \log
  p_\mathrm{x}(\vect{x})}{\partial \vect{x}}$ and, on average, carry
samples from areas of high density towards lower densities.  The
denoising function learns to oppose this, with the same force but
opposite sign\footnote{Notice the similarity to the denoising in
  Eq.~\eqref{eq:val-DSS-denoise}.}.  When sampling starts with any
given distribution, the combined steps of corruption and denoising
produce an average flow of samples which only disappears when the
diffusion flow caused by corruption exactly cancels the flow caused by
denoising, that is, when the sample distribution follows the original
training distribution.  \textcite{Bengio2013gsn} suggested that
sampling is more efficient in hierarchical models if corruption takes
place not only on inputs but on all levels of the encoder path and
called such networks generative stochastic networks (GSN).

What is surprising is that from denoising functions it is even
possible to derive probability estimates for the data.  Note that the
denoising function loses information about absolute probability and
only conserves information about relative probabilities because the
logarithm first turns multiplication into summation and the constant
normalization
term then disappears in differentiation.  Such a representation bears
similarity to energy-based probability models where only relative
probabilities can be readily accessed.  It turns out, however, that
any model which can reconstruct missing data can be turned into a
probability density estimator \parencite{Uria2014}.  By using input
erasure as corruption, the autoencoder can thus be used for deriving
normalized probability estimates even if denoising function loses
information about the normalization factor of the probability.

For our purposes, a particularly important feature of denoising
autoencoders is that they can handle over-complete representations,
including the shortcut connections in Eq.~\eqref{eq:val-shortcut}. The
reason for this is that it is not enough for the network to simply
copy its inputs to outputs since the inputs are corrupted by noise.
Rather, the network has to find a representation which makes removing
noise as easy as possible.

\subsection{Recursive derivation of the learning rule}

We are now ready to derive a learning rule with a distributed cost
function for the ladder network.  The basic idea is to apply a
denoising autoencoder recursively.  The starting point is the standard
denoising autoencoder which minimizes the following cost $C$:
\begin{eqnarray}
  \vect{\tilde{x}}(t) & = & \mathrm{corrupt}(\vect{x}(t)) \\
  \vect{\hat{x}}(t) & = & g(\vect{\tilde{x}}(t)) \\
  C & = & \frac{1}{T} \sum_{t = 1}^T ||\vect{x}(t) - \vect{\hat{x}}(t)||^2 \, .
\end{eqnarray}
During learning the denoising function $g$ learns to remove the
noise which we are injecting to corrupt $\vect{x}(t)$.  Now assume
that the denoising function uses some internal variables
$\vect{h}^{(1)}(t)$ for implementing a multi-layer mapping:
\begin{eqnarray}
  \vect{\tilde{h}}^{(1)}(t) & = & f^{(1)}(\vect{\tilde{x}}(t)) \\
  \vect{\hat{x}}(t) & = & g(\vect{\tilde{h}}^{(1)}(t)) \, .
\end{eqnarray}
Rather than giving all the responsibility of denoising to $g$, it is
possible to learn first how to denoise $\vect{h}^{(1)}$ and then use
that result for denoising $\vect{x}$:
\begin{eqnarray}
  \vect{h}^{(1)}(t) & = & f^{(1)}(\vect{x}(t)) \\
  \vect{\tilde{h}}^{(1)}(t) & = & f^{(1)}(\vect{\tilde{x}}(t)) \\
  \vect{\hat{h}}^{(1)}(t) & = & g^{(1)}(\vect{\tilde{h}}^{(1)}(t)) \\
  \vect{\hat{x}}(t) & = & g^{(0)}(\vect{\hat{h}}^{(1)}(t)) \\
  C^{(1)} & = & \frac{1}{T}
               \sum_{t = 1}^T ||\vect{h}^{(1)}(t) - \vect{\hat{h}}^{(1)}(t)||^2 \\
  C^{(0)} & = & \frac{1}{T}
               \sum_{t = 1}^T ||\vect{x}(t) - \vect{\hat{x}}(t)||^2 \, .
\end{eqnarray}
Training could alternate between training the mapping $g^{(1)}$ by
minimizing $C^{(1)}$ and training all the mappings by minimizing
$C^{(0)}$.

We can continue adding layers and also add the lateral connections of
the ladder network:
\begin{eqnarray}
  \vect{h}^{(l)}(t) & = & f^{(l)}(\vect{h}^{(l-1)}(t)) \\
  \vect{\tilde{h}}^{(l)}(t) & = & f^{(l)}(\vect{\tilde{h}}^{(l-1)}(t)) \\
  \vect{\hat{h}}^{(l)}(t) & = & g^{(l)}(\vect{\tilde{h}}^{(l)}(t),
                                   \vect{\hat{h}}^{(l+1)}(t)) \\
  C^{(l)} & = & \frac{1}{T}
               \sum_{t = 1}^T ||\vect{h}^{(l)}(t) - \vect{\hat{h}}^{(l)}(t)||^2 \, .
    \label{eq:val-Cl}
\end{eqnarray}
As before, we assume that $\vect{h}^{(0)}$ refers to the observations
$\vect{x}$.

This derivation suggests that the cost functions $C^{(l)}$ should only
be used for learning the encoding mappings $f^{(>l)}$ of the layers
above but not $f^{(\leq l)}$, those of the layers below, because
$C^{(l)}$ is derived assuming $\vect{h}^{(l)}(t)$ is fixed.  This is
problematic because it means that there cannot be a single consistent
cost function for the whole learning process and learning has to
continuously alternate between learning different layers.

However, recall that in the DSS framework it is precisely the forward
mapping $f^{(l)}$ which is updated using Eq.~\eqref{eq:val-C-DSS},
practically the same equation as Eq.~\eqref{eq:val-Cl}.  This means
that we can in fact minimize a single cost function
\begin{equation}
  C = C^{(0)} + \sum_{l = 1}^L \alpha_l C^{(l)} \, ,
  \label{eq:val-ladder-cost}
\end{equation}
where the coefficients $\alpha_l$ determine the relative weights of the
cost terms originating in different layers.

The DSS framework assumes that $\vect{\hat{h}}^{(l)}$ is constant and
optimizes using gradients stemming from $\vect{h}^{(l)}$ while in the
denoising autoencoder framework the roles are reversed.  This means
that by combining the cost functions and learning everything by
minimizing the cost with respect to all the parameters of the model,
we are essentially making use of both types of learning.

Just like in hierarchical latent variable models, higher level priors
offer guidance to lower-level forward mappings (cf.\ EM algorithm).
Since the gradients propagate backward along the encoding path, this
model is fully compatible with supervised learning: the standard
supervised cost function can simply be added to the top-most layer
$L$, measuring the distance between $\vect{h}^{(L)}(t)$ and the target
output.

\subsection{Decorrelation term for the cost function}

There is one final thing we must take care of: the decorrelation term
needed by DSS algorithms.  Recall that Eq.~\eqref{eq:val-Cl} is
minimized if $\vect{h}^{(l)}(t) = \vect{\hat{h}}^{(l)}(t) =$ constant.
Minimization of Eq.~\eqref{eq:val-Cl} with respect to
$\vect{\hat{h}}^{(l)}(t)$ actually typically promotes decorrelation
because it amounts to regression and any extra information can be used
to reduce the reconstruction error.  Minimization of
Eq.~\eqref{eq:val-Cl} with respect to $\vect{h}^{(l)}(t)$ promotes
finding projections that can be predicted as well as possible and,
since mutual information is symmetric, therefore also help predicting
other features as long as the entropy of the hidden unit activations
is kept from collapsing by avoiding the trivial solution where
$\vect{h}(t) =$ constant.

We are going to assume that the mappings $f^{(l)}$ and $g^{(l)}$ are
sufficiently general that we can, without loss of generality, assume
that the covariance matrix $\matr{\Sigma}^{(l)}$ of the hidden unit
activations on layer $l$ equals the unit matrix: $\matr{\Sigma}^{(l)} =
\matr{I}$, where
\begin{equation}
  \matr{\Sigma}^{(l)} = \frac{1}{T} \sum_{t = 1}^T 
     \vect{h}^{(l)}(t) [\vect{h}^{(l)}(t)]^T \, .
     \label{eq:val-Sigma}
\end{equation}
Here we assumed that the average activation is zero, a constraint
which we can also enforce without any loss of generality as long as the
first stages of mappings $f^{(l)}$ and $g^{(l)}$ are affine transformations.

A very simple cost function to promote $\matr{\Sigma}^{(l)} \approx \matr{I}$
would be $\sum_{i,j}[\Sigma_{ij}^{(l)} - \delta_{ij}]^2$, where
$\delta_{ij}$ is the Kronecker delta.  In other words, this
measures the sum of squares of the difference between $\matr{\Sigma}^{(l)}$
and $\matr{I}$.  However, this cost function does not distinguish
between too small and too large eigenvalues of
$\matr{\Sigma}^{(l)}$ but from the viewpoint of keeping the DSS-style
learning from collapsing the representation of $\vect{h}^{(l)}(t)$,
only too small eigenvalues pose a problem.  To analyse the situation,
note that
\begin{equation}
  \sum_{i,j}[\Sigma_{ij}^{(l)} - \delta(i, j)]^2 = 
    \tr \left( [\matr{\Sigma}^{(l)} - \matr{I}]^2 \right) =
    \sum_i (\lambda_i^{(l)} - 1)^2 \, ,
    \label{eq:val-lambda-1}
\end{equation}
where $\lambda_i^{(l)}$ are the eigenvalues of $\matr{\Sigma}^{(l)}$.
The first equality follows from the definition of the trace of a
matrix and the second from the fact that the trace equals the sum of
eigenvalues.

Since Eq.~\eqref{eq:val-lambda-1} is symmetric about $\lambda = 1$,
it penalizes $\lambda = 0$ just as much as $\lambda = 2$ while the
former is infinitely worse from the viewpoint of keeping $\vect{h}$
from collapsing.

A sound measure for the information content of a
variable is the determinant of the covariance matrix because it
measures the square of the (hyper)volume of the (hyper)cuboid whose
sides have the length determined by the standard deviations of the
distribution along its eigenvectors. Since the determinant of a matrix
equals the product of its eigenvalues, the logarithm of the
determinant equals the sum of the logarithms of the eigenvalues:
\begin{equation}
  \log \det \matr{\Sigma}^{(l)} = \sum_i \log \lambda_i^{(l)} =
    \tr \left( \log \matr{\Sigma}^{(l)} \right) \, .
    \label{eq:val-lambda-2}
\end{equation}
The latter equality follows from the fact that any analytical function
can be defined for square matrices so that it applies to the eigenvalues of
the matrix. This is because $\left( \matr{E} \matr{\Lambda}
\matr{E}^{-1} \right)^k = \matr{E} \matr{\Lambda}^k \matr{E}^{-1}$ and
therefore any power series expansion of a matrix turns into the same
power series expansion of the eigenvalues.
Note that $\log \Sigma$ is the matrix logarithm, \textit{not} the logarithm
of the elements of the matrix.

Equation~\eqref{eq:val-lambda-2} is a measure which grows smaller when
the information content diminishes but it can be turned into a sensible
cost function which reaches its minimum value of zero when $\lambda
= 1$:
\begin{equation}
  C_{\Sigma}^{(l)} = \sum_i (\lambda_i^{(l)} - \log \lambda_i^{(l)} - 1) =
    \tr \left(\matr{\Sigma}^{(l)} - \log \matr{\Sigma}^{(l)} - \matr{I} \right)
    \label{eq:val-C-Sigma}
\end{equation}
This cost penalizes $\lambda = 0$ infinitely and grows relatively
modestly for $\lambda_i > 1$. 

It is relatively simple to differentiate this cost with
respect to $\matr{\Sigma}^{(l)}$ since, for any
analytical function $\phi$, it holds
\begin{equation}
  \frac{\partial \tr \left( \phi(\matr{\Sigma}) \right)}{\partial \matr{\Sigma}}
  = \phi'(\Sigma) \, .
\end{equation}
In our case $\phi(a) = a - \log a - 1$ and thus $\phi'(a) = 1 - a^{-1}$.
We therefore have
\begin{equation}
  \frac{\partial C_{\Sigma}^{(l)}}{\partial \matr{\Sigma}^{(l)}} = 
  \matr{I} - [\matr{\Sigma}^{(l)}]^{-1} \, .
\end{equation}
The rest of the formulas required for computing the gradients
with the chain rule are
straight-forward since Eq.~\eqref{eq:val-Sigma} has a simple quadratic form.

Note that all twice differentiable cost functions that are minimized
when $\lambda_i = 1$ have the same second-order behaviour (up to
scaling) close to the minimum so the simpler
Eq.~\eqref{eq:val-lambda-1} works just as well if all $\lambda_i$ are
sufficiently close to $1$.  However, to avoid any potential problems,
Eq.~\eqref{eq:val-C-Sigma} was used in the experiments presented in
this chapter.

Finally, as suggested earlier, we will add a simple term to the cost
function to make sure that the hidden unit activations really have a
zero mean:
\begin{eqnarray}
  \vect{\mu}^{(l)} & = & \frac{1}{T} \sum_{t = 1}^T  \vect{h}^{(l)}(t) \\
  C_{\mu}^{(l)} & = & ||\vect{\mu}^{(l)}||^2 \, .
\end{eqnarray}

While DSS algorithms require decorrelation of the output
representation, it is now also important to decorrelate (i.e., whiten)
the inputs.  Normally denoising autoencoders have a fixed input but
now the cost functions on the higher layers can influence their input
mappings and this creates a bias towards PCA-type solutions.  This is
because the amount of noise injected to $\vect{x}$ is relatively
smaller for projections for which the variance of $\vect{x}$ is
larger.  The terms $C^{(\geq 1)}$ are therefore smaller if the network
extracts mainly projections with larger variance, that is, PCA-type
solutions.  While PCA may be a desirable in some cases, often it is
not.

\subsection{Learning rule for the ladder network}

We are now ready to collect together the recipe for learning
the ladder network. Given (typically pre-whitened)
observations $\vect{h}^{(0)}(t) :=
\vect{x}(t)$, the cost function $C$ is computed using the following
formulas:
\begin{eqnarray}
  \label{eq:val-ladder-1}
  \vect{h}^{(l)}(t) & = & f^{(l)}(\vect{h}^{(l-1)}(t))
                      \mbox{\hspace{2.085cm} for $1 \leq l \leq L$} \\
  \vect{\tilde{h}}^{(0)}(t) & = & \mathrm{corrupt}(\vect{h}^{(0)}(t)) \\
  \label{eq:val-corrupt-h}
  \vect{\tilde{h}}^{(l)}(t) & = & f^{(l)}(\vect{\tilde{h}}^{(l-1)}(t))
                      \mbox{\hspace{2.085cm} for $1 \leq l \leq L$} \\
  \vect{\hat{h}}^{(L)}(t) & = & g^{(L)}(\vect{\tilde{h}}^{(L)}(t)) \\
  \vect{\hat{h}}^{(l)}(t) & = & g^{(l)}(\vect{\tilde{h}}^{(l)}(t),
    \vect{\hat{h}}^{(l+1)}(t)) \mbox{\hspace{1cm} for $0 \leq l \leq L-1$} \\
  C^{(l)} & = & \frac{1}{T}
               \sum_{t = 1}^T ||\vect{h}^{(l)}(t) - \vect{\hat{h}}^{(l)}(t)||^2 \\
  \matr{\Sigma}^{(l)} & = & \frac{1}{T} \sum_{t = 1}^T 
     \vect{h}^{(l)}(t) [\vect{h}^{(l)}(t)]^T \\
  C_{\Sigma}^{(l)} & = &
  \tr \left(\matr{\Sigma}^{(l)} - \log \matr{\Sigma}^{(l)} - \matr{I} \right) \\
  \vect{\mu}^{(l)} & = & \frac{1}{T} \sum_{t = 1}^T  \vect{h}^{(l)}(t) \\
  C_{\mu}^{(l)} & = & ||\vect{\mu}^{(l)}||^2 \\
  C & = & C^{(0)} + \sum_{l = 1}^L \alpha_l C^{(l)} + \beta_l C_{\Sigma}^{(l)} +
                                \gamma_l C_{\mu}^{(l)} \, .
  \label{eq:val-ladder-last}
\end{eqnarray}
Learning the parameters of the mappings $f^{(l)}$ and $g^{(l)}$ is
based on minimizing $C$.  The simple solution is to apply gradient
descent (stochastic or batch version) but basically any optimization
method can be used, for example nonlinear conjugate gradient or
quasi-Newton methods.  Whichever method is chosen, the existence of a
single cost function to be minimized guarantees that learning
converges as long as the minimization method performs properly.

Equation~\eqref{eq:val-corrupt-h} could include corruption in the
same manner as in GSN. However, to keep things simple, the experiments
reported in this chapter only applied corruption to the input layer.

\begin{figure}
  \centering
  \hspace*{-8mm}\input{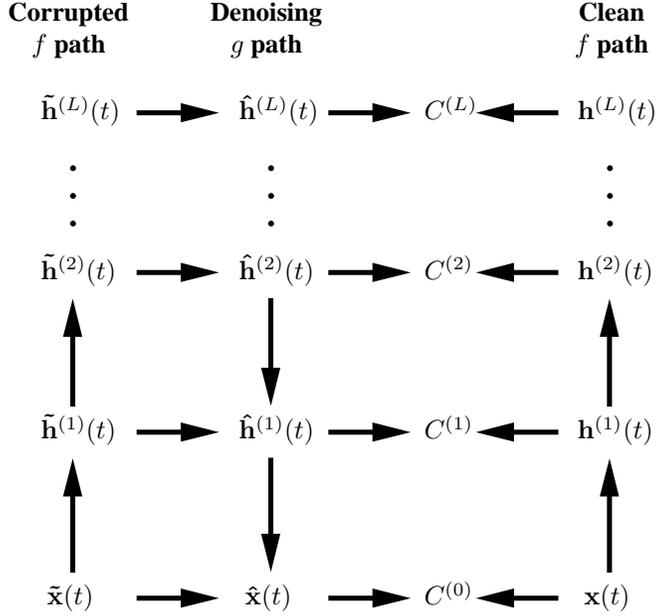} 
  \caption{Ladder network's cost computations are illustrated.  The
    clean $f$ path shares exactly the same mappings $f^{(l)}$ as the
    corrupted $f$ path.  The only difference is that corruption noise
    is added in the corrupted path.  The resulting corrupted
    activations are denoted by $\vect{\tilde{h}}^{(l)}(t)$.  On each
    layer, the cost function has a term $C^{(l)}$ which measures the distance
    between the clean activations $\vect{h}^{(l)}(t)$ and their
    reconstructions $\vect{\hat{h}}^{(l)}(t)$.  The terms
    $C_{\Sigma}^{(l)}$ and $C_{\mu}^{(l)}$ which measure how well the
    activations $\vect{\hat{h}}^{(l)}(t)$ are normalized are not
    shown.}
  \label{fig:val-ladder-cost}
\end{figure}

Figure~\ref{fig:val-ladder-cost} depicts the computational diagram of
the cost function of the ladder network.  The two $f$ paths going
upward share their mappings $f^{(l)}$ and the only difference is that
the inputs on the corrupted path are corrupted by noise.  Each layer
of the network adds its own term to the cost function, measuring how
well the clean activations $\vect{h}^{(l)}(t)$ are reconstructed from
the corrupted activations.  During forward computations, information
will flow from the observations towards the cost function terms along
the arrows.  During learning, gradients will flow from
the cost function terms in the opposite direction.  Training signals
arriving along the clean $f$ path correspond to DSS-style learning
while the training signals in the denoising $g$ path and corrupted $f$
path correspond to the type of learning taking place in denoising
autoencoders.  The terms $C_\Sigma^{(l)}$ and $C_\mu^{(l)}$ which
promote unit covariance and zero mean of the clean activations
$\vect{h}^{(l)}(t)$, respectively, are not shown.  They are required
for keeping DSS-style learning from collapsing the representations and
are functions of the clean $f$ path only.

From the perspective of learning deep networks, it is important that
any mapping, $f^{(l)}$ or $g^{(l)}$, is close to one of the cost
function terms $C^{(l)}$.  This means that learning is efficient even
if propagating gradients through the mappings would not be efficient.

\section{Experiments}
\label{sec:val-experiments}

This section presents a few simple experiments which demonstrate the
key aspects of the ladder network:
\begin{itemize}
\item How denoising functions can represent probability distributions.
\item How lateral connections relieve the pressure to represent
  every detail at the higher layers of the network and allow them to
  focus on abstract invariant features.
\item How the cost function terms on higher layers speed up learning.
\end{itemize}
The three following sections will gradually develop a two-layered
ladder network which can learn abstract invariant features.  First,
simple distributions are modeled.  Then this is put into use in a
linear ICA model with one hidden layer.  Finally, a second layer is
added which models the correlations between the variances of the first
layer activations. All the experiments used the set of learning rules
described in Eqs.~(\ref{eq:val-ladder-1}--\ref{eq:val-ladder-last}).
The hyperparameters $\beta_l$ were automatically adjusted to keep the
smallest eigenvalue of $\matr{\Sigma}_l$ above $0.7$.  The
hyperparameter $\gamma_l$ was set to the the same value as $\beta_l$.

\subsection{Representing distributions with denoising functions}

\begin{figure}
  \centering
  \input{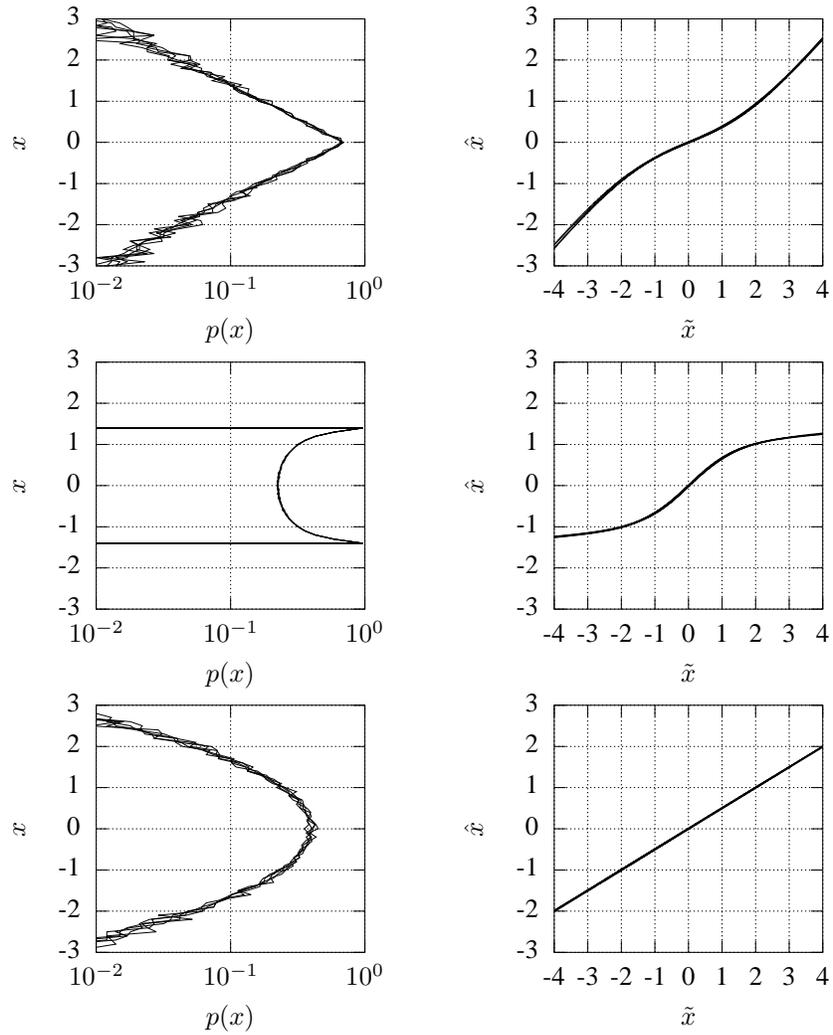}
  \caption{Illustration of the connection between the marginal
    distribution and denoising function. On the left, three different
    probability distributions are shown on a logarithmic scale.
    From top: super-Gaussian, sub-Gaussian and Gaussian distribution.
    Each distribution has a unit variance and zero mean.
    On the right, denoising functions have been been trained to
    remove corruptive noise with unit variance. In the Gaussian case,
    theoretically the optimal solution is $\hat{x} = \tilde{x} / 2$.
    Each plot shows five different random samples plotted on top of
    each other.  Note that in the plots showing the pdf or the data,
    $x$ is plotted on the vertical axis to help side-by-side comparison
    with the denoising function.}
  \label{fig:val-denoising}
\end{figure}

We will start by a simple experiment which elucidates the relation
between the prior distribution of activations and their denoising
functions, illustrated in Fig.~\ref{fig:val-denoising}.  Three
different distributions were tested, super-Gaussian, sub-Gaussian and
Gaussian distribution. Each of them had a unit variance and zero mean.
Each plot shows five different results overlaid on top of each other.

The super-Gaussian distribution was a Laplace distribution whose
probability density function (pdf) is
$$
  p(x) = \frac{1}{\sqrt{2}} e^{-\sqrt{2}|x|} \, .
$$
On a logarithmic scale, it behaves as $-\sqrt{2} |x|$ plus constant.

The sub-Gaussian distribution was generated by scaling a sinusoidal
signal by $\sqrt{2}$ to obtain a variable with unit variance.  The pdf
of this distribution is
$$
  p(x) = \frac{1}{\pi \sqrt{2 - x^2}}
$$
for $|x| < \sqrt{2}$.

For this experiment, the model did not have any hidden layers ($L =
0$) and therefore there are no forward functions $f$, only one
denoising function which was implemented as a single hidden neuron
with tanh activation and a bypass connection:
$$
  \hat{x} = g(\tilde{x}) = \xi_1 \tilde{x} +
    \xi_2 \tanh (\xi_3 \tilde{x} + \xi_4) + \xi_5 \, ,
$$
where $\xi_i$ are scalar parameters.  All parameters
were optimized by minimizing $C^{(0)} = \sum_t ||\hat{x}(t) - x(t)||^2$.

With small enough noise, the denoising functions should theoretically
approach
$$
  \hat{x} \approx \tilde{x} + \sigma_{\mathrm{n}}^2
    \frac{\partial \log p(x)}{x}_{x = \tilde{x}} \, ,
$$
where $\sigma_{\mathrm{n}}^2$ is the variance of the noise used for
corrupting $\tilde{x}$.  With small corruption noise, this would then
mean that the denoising function of the Laplacian input would be a sum
of $x$ and a scaled step function.  This is not the case now since
$\sigma_{\mathrm{n}}^2$ is as large as the variance of the input.  This
tends to smoothen the denoising function.

As can be readily seen from Fig.~\ref{fig:val-denoising}, the
denoising function for a Gaussian observation is linear. Theoretically,
the function should be
$$
  \hat{x} = \frac{\sigma_{\mathrm{x}}^2}{\sigma_{\mathrm{x}}^2 + \sigma_{\mathrm{n}}^2}
     \tilde{x} \, ,
$$
where $\sigma_{\mathrm{x}}^2$ is the variance of the observations.
Since both variances equal one in these experiments, the theoretical
optimum is $\hat{x} = \tilde{x} / 2$.  The estimated denoising
function follows this very closely.

\subsection{ICA model}

We will now move on to a simple linear ICA model which serves as an
example of how statistical modeling will be translated into function
approximation in this framework.  It also gives some intuition on
how the lateral connections help higher levels to focus on relevant
features.

In linear independent component analysis, the data is assumed to be a
linear mixture of independent identically distributed sources. Unlike
in principal component analysis, the mixing is not restricted to be
orthogonal because sources with non-Gaussian marginal distributions
can be recovered.  If more than one source has a Gaussian
distribution, these sources cannot be expected to be recovered
and will remain mixed. However, the subspace spanned by the Gaussian
sources should be recoverable and separate from all the non-Gaussian sources.

One limitation is that unless there is some extra information
available, the sources can only be recovered up to scaling and
permutation.  Scaling is usually fixed by assuming that source
distributions have a unit variance.  This will still leave permutation
and sign of the sources ambiguous.

The dataset was generated by linearly mixing 10,000 samples from 15
sources into 15 observations.  The elements of the mixing matrix were
sampled from a zero-mean Gaussian distribution.  The souces had the
same distributions as in the previous example: five super-Gaussian,
five sub-Gaussian and five Gaussian sources.  The data were
\textit{not} whitened because one purpose of the experiments was
to demonstrate that normal autoencoders have a bias towards PCA solution
even if the cost function terms $C^{(\geq 1)}$ are not used.  In these
experiments, $\alpha_l$ in Eq.~\eqref{eq:val-ladder-last} were set to zero.

\subsubsection{Model structure}

The model had one hidden layer ($L = 1$) and the only nonlinearity was
on the hidden layer denoising which was a simplified version of the
model used in the previous experiment.  With essentially zero mean
observations, there is no need for bias terms in the model.  The mappings
of the model are as follows:
\begin{eqnarray}
  \label{eq:val-ica-1}
  f(\vect{x}) & = & \matr{W} \vect{x} \\
  \label{eq:val-ica-2}
  g^{(1)}_i(\vect{h}) & = & a_i h_i + b_i \tanh(h_i) \\
  g^{(0)}(\vect{x}, \vect{h}) & = & \matr{A} \vect{h} + \matr{B} \vect{x}
  \label{eq:val-ica-3}
\end{eqnarray}
The parameters to be estimated are the three matrices $\matr{W}$,
$\matr{A}$ and $\matr{B}$ and the two vectors $\vect{a}$ and
$\vect{b}$ which were used for the denoising functions of individual
hidden units.

Note that the underlying assumption of ICA models is that the sources
are independent.  This is incorporated in the model by making the
denoising on the hidden layer unit-wise.  Also, the lateral linear
mapping with matrix $\matr{B}$ can model any covariance structure in
the observation data.  This means that the hidden layer should be able
to focus on representing the non-Gaussian sources.

\subsubsection{Results}

Experiments verify that the model structure defined by
Eqs.~(\ref{eq:val-ica-1}--\ref{eq:val-ica-3}) is indeed able to
recover the original sources that were used for generating the
observed mixtures (up to permutation and scaling).  This is apparent
when studying the normalized loading matrix which measures how much
the original sources contribute the the values of the hidden units.
The loading matrix is obtained from the product
$\matr{W} \matr{A}_{\mathrm{orig}}$ where
$\matr{W}$ is the unmixing matrix learned by the model and
$\matr{A}_{\mathrm{orig}}$ is the original mixing matrix.
Rows of this matrix measure how much
contribution from the original source there is in each of the
recovered hidden neuron activation.  In the normalized loading matrix,
the rows are scaled so that the squares sum up to one, that is, the row
vectors have unit lengths.
A successful unmixing is
characterized by a single dominant loading in each row and can be
measured by the average angle of each vector to the dominant source.
In the experiments, a typical value was around $10\textdegree$ which
corresponds to the contribution $\cos (10\textdegree) \approx 0.985$ from the
dominant source.

The denoising mappings $g^{(1)}_i(\vect{h})$ for each source depend on
the distribution of the source just as expected.  In particular, the
sign of the parameter $b_i$ is determined by the super- or
sub-Gaussianity of the sources.  When there are more hidden units than
non-Gaussian sources, the model will also represent Gaussian sources
but the preference is for non-Gaussian sources.  This is to be
expected because the lateral mapping $\matr{B}$ at the lowest level of
the network can already represent any Gaussian structure.  In other
words, the model performs just as expected.

What is more interesting is what happens if the lateral mapping
$\matr{B}$ is missing.  Since the reconstruction $\vect{\hat{x}}(t)$
can then only contain information which is present in the hidden
units, the network has a strong pressure to conserve as much
information as possible.  Essentially the dominant mode of operation
is then PCA: the network primarily extracts the subspace spanned by
the eigenvectors of the data covariance matrix corresponding to the
largest eigenvalues.  The network can only secondarily align the
representation along independent components.  If the independent
components do not happen to align with the principal subspace, PCA
wins over ICA.

In one experiment, for instance, a network with $\matr{B}$ and 11
hidden units was able to retrieve the ten non-Gaussian sources with
loadings between 0.958 and 0.994, averaging 0.981.  By contrast, the
ten best loadings in exactly the same setting but without $\matr{B}$
were between 0.499 and 0.824 and averaged 0.619 after the same number
of iterations.  This is not significantly different from
random mappings which yield average loadings around
$0.613 \pm 0.026$ (average $\pm$ std).

It turned out that the network was able to do better but it convereged
tremendously slowly, requiring about 100 times more iterations than
with $\matr{B}$.  Still, even after the network had seemingly converged,
the ten best loadings were between 0.645 and 0.956 and averaged 0.870.
While this is clearly better than random, even the best loading was worse
than the worst loading when the lateral connections $\matr{B}$ were used.

That this is due to the
network's tendency to extract a principal subspace can be seen by
analysing how large portion of the subspace spanned by $\matr{W}$
falls outside the subspace spanned by the 11 largest eigenvectors of
the data covariance matrix.  In the network with $\matr{B}$ this was
28~\% whereas in the network lacking $\matr{B}$ this was just 0.03~\%.

To be fair, it should be noted that pre-whitening the inputs
$\vect{x}(t)$ restores the autoencoder's ability to recover
independent components just as it allows nonlinear PCA learning rule
to perform independent component analysis rather than principal
subspace analysis.  However, in more complex cases it may not be as
easy to normalize away the information that is not wanted.

\subsection{Hierarchical variance model}

We shall now move on to expanding the ICA model by adding a new layer
to capture the nonlinear dependencies remaining in the hidden unit
activations in the ICA model.  This makes sense because it is usually
impossible to produce truly statistically independent components
simply by computing different linear projections of the observations.
Even if the resulting feature activations lack linear correlations,
there are normally higher-order dependencies between the features
\parencite[for more discussion of such models,
  see][]{Valpola04SigProc,HyvHoy00NC}.

One typical example is that the variances of the activations are
correlated because the underlying cause of a feature activation is
likely to generate other activations, too.  In order to find a network
structure that could represent such correlated variances, let us
recall that the optimal denoising of a Gaussian variable $h$ with
prior variance $\sigma_{\mathrm{h}}$ and Gaussian corruption noise
$\sigma_{\mathrm{n}}$ is
$$
  \hat{h} = \frac{\sigma_{\mathrm{h}}^2}{\sigma_{\mathrm{h}}^2 + \sigma_{\mathrm{n}}^2}
     \tilde{h} \, .
$$
This can be written as
$$
  \hat{h} = \frac{1}{1 + \sigma_{\mathrm{n}}^2 / \sigma_{\mathrm{h}}^2} \tilde{h} =
  \sigmoid(\log \sigma_{\mathrm{h}}^2 - \log \sigma_{\mathrm{n}}^2) \tilde{h} \, ,
$$
where the sigmoidal activation function is defined as
$$
  \sigmoid(x) = \frac{1}{1 + e^{-x}} \, .
$$
What this means is that information about the variance of $h$
translates to a modulation of the connection strength in the denoising
function mapping $\tilde{h}$ to $\hat{h}$.

The experiments in this section demonstrate that all we need to do to
represent correlations in the variances in the hidden unit activations
is add modulatory connections from the layer above and give enough
flexibility for the forward mapping to the highest layer.  The network
will then find a way to use the representations of the highest layer.
Moreover,
lateral connections on the middle layer now play a crucial role
because they allow the higher layers to focus on representing
higher-order correlations.  Unlike in the case with the ICA model
in the previous section, this could not have been replaced by whitening
the inputs.  The experiments also demonstrate how the
cost function terms on the higher layers speed up learning.

\subsubsection{Data}

The dataset of 10,000 samples was generated as a random linear mixture
of sources $s_i$ just as in the ICA experiment. This time, however, the
sources were Gaussian with a changing variance.  The variances of the
sources were determined by higher-order variance sources, each of
which was used by a group of four sources. There were four such
groups, in other words, four higher-order variance sources which
determined the variances of 16 sources.  Such groups of dependent
sources mean that the data follows the model used in independent
subspace analysis \parencite{HyvHoy00NC}.

The variance sources $v_j$ were sampled from a Gaussian distribution and the
variance $\sigma_i^2$ of the lower-level sources $i \in \set{G}_j$
was obtained by computing $e^{v_j}$.  The set $\set{G}_j$ contain all
the indices $i$ for which the lower-level sources $s_i$ are are modulated by
the variance source $v_j$.  Since there were four non-overlapping groups
of four sources, $\set{G}_1 = \{1, 2, 3, 4\}$, $\set{G}_2 = \{5, 6, 7, 8\}$,
and so on.

Note that although the sources are sampled from a Gaussian
distribution, their marginal distribution is super-Gaussian since the
variance is changing.  In these experiments, the dataset was
pre-whitened.

\subsubsection{Model structure}

The linear mappings between the observations and the first layer were
the same as with the ICA model but now a second nonlinear layer was
added and denoising $g^{(1)}$ of the first layer was modified to make use of
it:
\begin{eqnarray}
  \label{eq:val-var-f1}
  f^{(1)}(\vect{x}) & = & \matr{W}^{(1)} \vect{x} \\
  \label{eq:val-var-f2}
  f^{(2)}(\vect{h}^{(1)}) & = & \mathrm{MLP}(\vect{h}^{(1)}) \\
  \label{eq:val-var-g2}
  g^{(2)}_i(\vect{h}^{(2)}) & = & a_i h^{(2)}_i \\
  \label{eq:val-var-g1}
  g^{(1)}_i(\vect{h}^{(1)}, \vect{h}^{(2)}) & = & 
    \sigmoid(\matr{A}^{(2)}_i \vect{h}^{(2)} + b^{(1)}_i) h^{(1)}_i \\
  \label{eq:val-var-g0}
  g^{(0)}(\vect{x}, \vect{h}^{(1)}) & = &
    \matr{A}^{(0)} \vect{h}^{(1)} + \matr{B} \vect{x} \, .
\end{eqnarray}
The multi-layer perceptron (MLP) network used in the second-layer
encoder mapping $f^{(2)}$ was
\begin{equation}
  \mathrm{MLP}(\vect{h}) = \matr{W}^{(2b)} \psi(\matr{W}^{(2a)} \vect{h}
    + \vect{b}^{(2a)}) + \vect{b}^{(2b)} \, ,
\end{equation}
where the activation function $\psi(x) = \log(1 + e^x)$ operates on
the elements of the
vector separately.  Note that we have included the bias term
$\vect{b}^{(2b)}$ to make sure that the network can satisfy the
constraint of having zero mean activitions.

\subsubsection{Results}

\begin{figure}
  \centering
  \input{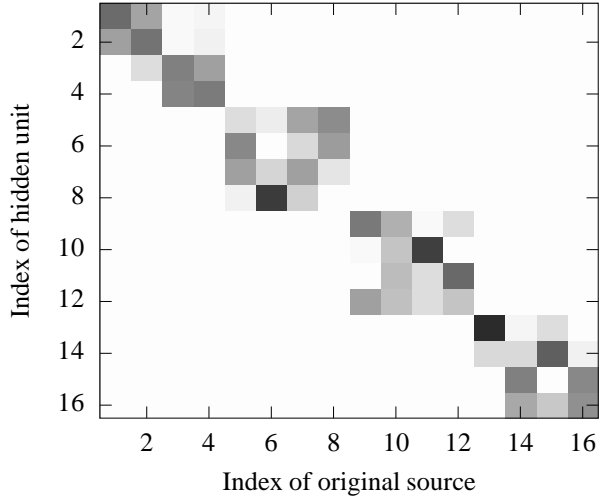}
  \caption{The squares of the normalized loading matrix from a
    hierarchical ladder network.  Black corresponds to
    one and white to zero.  As is evident from
    the plot, the four subspace $\set{G}_j$ have been cleanly separated from
    each other but remain internally mixed.  This is because
    the distribution of the sources within each subspace
    is spherically symmetric which makes it impossible
    to determine the rotation within the subspace. This is reflected
    in the blocky structure of the matrix of loadings.
    The hidden units were ordered appropriately to reveal
    the block structure.}
  \label{fig:val-var-mixing}
\end{figure}

\begin{figure}
  \centering
  \hspace*{-1cm}\input{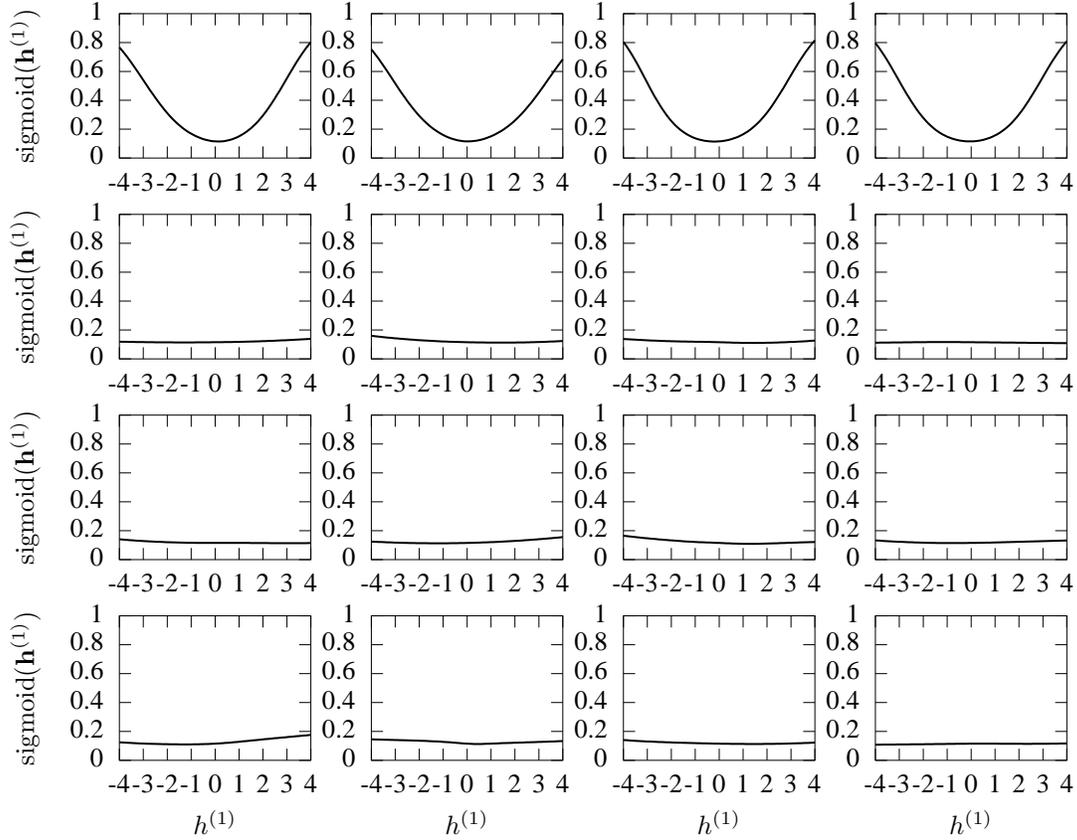}
  \caption{Illustration of one of the denoising functions learned by a
    hierarchical ladder network. Each plot shows how the term
    $\sigmoid(\matr{A}^{(2)}_1 \vect{h}^{(2)} + b^{(1)}_1)$ modulating
    $\tilde{h}^{(1)}_1$ in
    Eq.~\eqref{eq:val-var-g1} behaves as a function of one of the
    hidden neurons $h^{(1)}_i$. In each plot, $h^{(1)}_i$ for just one
    $i$ takes nonzero values. The top row shows how $\sigmoid$
    changes as a function of the hidden units which belong to the same
    subspace $\set{G}_1$ as $h^{(1)}_1$ whose sigmoid is shown.
    The rest of
    the plots correspond to hidden units from other groups $\set{G}_{\neq 1}$.}
  \label{fig:val-var-denoising}
\end{figure}

Experiments verified that the network managed to separate individual
source subspaces (Fig.~\ref{fig:val-var-mixing}) and learned to model
the correlations between the variances of different sources
(Fig.~\ref{fig:val-var-denoising}).  The figures correspond to an
experiment where the dimension of the first layer was 16 and second
layer 10. The MLP network used for modeling $f^{(2)}$ had 50 hidden
units. The figures show results after 1000 training for iterations.
The results were relatively good already after 300 iterations and
after 1000 iterations the network had practically converged.

Figure~\ref{fig:val-var-mixing} shows squares of the loadings, that is,
elementwise squares of the matrix $\matr{W}^{(1)}
\matr{A}_{\mathrm{orig}}$, scaled such that the squares sum up to one
for each hidden unit.  Since the data was generated by modulating the
variance of the Gaussian sources of each subspace, the data
distribution was spherically symmetric within each subspace. This is
reflected in the blocky appearance of the the matrix of loadings as
the network has settled on a random rotation within the subspace. Note
that such rotation indeterminacy is a property of the input data. In
other experiments the model was fed with the ICA data from the
previous section and readily separated the sources (results not shown
here).

Figure~\ref{fig:val-var-denoising} depicts the learned denoising
functions.  Each plot shows how the term
$\sigmoid(\matr{A}^{(2)}_1 \vect{h}^{(2)} + b^{(1)}_1)$ belonging to
the first hidden unit in Eq.~\eqref{eq:val-var-g1} behaves as a
function of one of the hidden neurons $h^{(1)}_i$. The sigmoid term is
a function of all the hidden neuron activations through the second
layer hidden unit activations.  Theoretically, the sigmoid term
modulating the mapping from $\tilde{h}^{(1)}_1$ to $\hat{h}^{(1)}_1$ should
be a function of the norm of the activation vector of subspace $\set{G}_1$,
that is, a function of $\sum_{i \in \set{G}_1} [h^{(1)}_i]^2$.
As is readily seen from the plots, the
second layer activations have apparently learned to represent such
quadratic terms of first layer activations because the sigmoid indeed
appears to be a function of the norm of the activation vector of the
first subspace.  When ICA data was used (results not shown here), the
sigmoid term learned to neglect all the other hidden units and
developed to be a function of $h_1^{(1)}$ only (assuming we are
looking at the sigmoid of hidden unit 1).

The variance features developing on the second layer of the model can
only improve the reconstruction of $\hat{\vect{x}}$ if they are
combined with $\vect{h}^{(1)}$ because variance alone cannot say
anything about the direction where the reconstruction should be
changed.  Without the shortcut connections in the ladder model, the
highest layer with only ten hidden units\footnote{In principle the
  highest layer only needs four hidden units to represent the four
  variance sources.  The network was indeed able to learn such a
  compact representation but learning tended to be slower than with
  ten hidden units.} could not have learned to
represent the variance sources because there would not have been
enough space to also represent the activations $\vect{h}^{(1)}$ which
are also needed to make use of the variance sources. The higher layers
can only let go of the details because they are recovered again when
denoising proceeds from the highest layers towards the lowest.

Another important question was whether the terms of the cost function
originating in the higher layers of the network really are useful for
learning.  To investigate this, 100 simulations were run with
different datasets and random initializations of the network.  It
turned out that in this particular model, the higher layer cost
function term $C^{(2)}$ was not important but $C^{(1)}$ could speed up
learning considerably particularly during the early stages of
learning.  As expected, it was crucial to combine it with a proper
decorrelation term $C_{\Sigma}^{(1)}$.  The success of the model was
measured by the value $C^{(0)}$ that was reached.  Note that it is not
\textit{a priori} clear that adding other cost function terms could
help reduce $C^{(0)}$. Nevertheless, this turned out to be the case.
By iteration 100, the network consistently reached a lower value of
$C^{(0)}$ than by iteration 200 when minimizing $C^{(0)}$ alone (as in
standard denoising autoencoders). Subsequent learning continued
approximately at the same pace which seems reasonable as denoising
autoencoders should be able to optimize the model after it has been
initialized close to a sensible solution.

Another interesting finding was that about one third of the
improvement seems to be attributable to the decorrelation term
$C_{\Sigma}^{(1)}$. It was able to speed up learning alone without
$C^{(1)}$ despite initializing the network with mappings which
ensure that all the representations start out as decorrelated. Whereas
the speedup of $C^{(1)}$ was most pronounced in the beginning of
learning, the speedup offered by $C_{\Sigma}^{(1)}$ was more important
during the middle phases of learning. Presumably this is because the
representations start diverging from the uncorrelated initialization
gradually.

At the optimum of the cost function $C^{(0)}$, the addition of any
extra term can only make the situation worse from the viewpoint of
minimizing $C^{(0)}$. It is therefore likely that optimally the
weights $\alpha_l$ and $\beta_l$ in Eq.~\eqref{eq:val-ladder-last}
should be gradually decreased throughout training and could be set to
zero for a final finetuning phase.  For simplicity, all $\alpha_l$
were kept fixed in the simulations presented here.

\section{Discussion}
\label{sec:val-discussion}

The experiments verified that the ladder model with lateral shortcut
connections and cost function terms at every level of the hierarchy is
indeed able to learn abstract invariant features efficiently.
Although the networks studied here only had a few layers (no more than
six between $\vect{\tilde{x}}$ and $\vect{\hat{x}}$ no matter how they
are counted), what is important is that the representations were
abstract and invariant already on the second layer.  In fact, the
model with two layers, linear features on the first and variance
features on the second, corresponds roughly to the architecture with
simple and complex cells found by \textcite{Hubel:62} \parencite[for a
  more detailed discussion, see][]{HyvHoy00NC}. Promising as the
results are, it is clearly necessary to conduct far larger experiments
to verify that the ladder network really does support learning in much
deeper hierarchies.

Similarly, it will be important to verify that the ladder network is
indeed compatible with supervised learning and can therefore support
useful semi-supervised learning.  All the experiments reported in this
chapter were unsupervised. However, all the results supported the
notion that the shortcut connections of the ladder network allow it to
discard information, making it a good fit with supervised learning.

The approach proposed here is not the only option for semi-supervised
learning with autoencoders. Another alternative is to split the
representation on the higher levels to features which are also used by
supervised learning and to features which are only used by the decoder
\parencite{Rifai2012}. Recently \textcite{Kingma2014NIPS} obtained
good results with a model which combined this idea with hierarchical
latent variable models.  Their model included a trainable recognition
mapping in addition to the generative model, so the overall structure
resembles an autoencoder.  Only additive interactions were included
between the encoder and decoder and it remains to be seen whether their
approach can be extended to include more complex interactions, such as
the modulatory interaction in Eq.~\eqref{eq:val-var-g1}.

A somewhat similar approach, motivated by variational autoencoders and
target propagation, was proposed by \textcite{Bengio2014TP}. It also
includes exchange of information between the decoder and encoder
mappings on every level of the hierarchy, though only during learning.

One of the most appealing features of the approach taken here is that it
replaces all probabilistic modeling with function approximation
using a very simple cost function.  In
these experiments, an MLP network learned to extract higher-level
sources that captured the dependencies in the first-level sources.
This model corresponded to independent subspace analysis not because
the model was tailored for it but because the input data had that
structure.  The forward mapping $f^{(2)}$ was very general, an MLP
network.  In these experiments the denoising mapping $g^{(1)}$ had a
somewhat more limited structure mainly to simplify analysis of the
results.  It will be interesting to study whether $g^{(l)}$ can also
be replaced by a more general mapping.

Another important avenue for research will be to take advantage of all the
machinery developed for GSN and related methods, such as sampling, calculating
probability densities and making use of multiple rounds of corruption
and denoising during learning
\parencite{Bengio2013gsn,Uria2014,raiko2014iterative}.
Particularly the ability to
sample from the model should be very useful. In order to make full use
of these possibilities, the corruption procedure should be
extended. Now simple Gaussian noise was added to the inputs. Noise
could be added at every layer to better support sampling
\parencite{Bengio2013gsn} and could also involve masking out some
elements of the input vector completely \parencite{Uria2014}.  If
different types of corruption are needed at different times, it might
be possible to extend the denoising functions to handle different
types of corruption (information about the corruption strategy could
be provided as side information to the denoising functions) or it
might be possible to relearn just the denoising functions $g^{(l)}$
while keeping the previously learned forward mappings $f^{(l)}$ fixed.

A crucial aspect of the ladder network is that it captures the
essential features of the inference structure of hierarchical latent
variable models. Along the same line, it should be possible to extend
the model to support even more complex inferences, such as those
that take place in Kalman filters.

The model studied by \textcite{Yli-Krekola:2007} takes even one step
further: it implements a dynamical biasing process which gives rise to
an emergent attention-like selection of information in a similar
fashion as in the model suggested by \textcite{decojarolls}.  The
model studied by \textcite{Yli-Krekola:2007}
is derived from the DSS framework and
already has a structure which is reminiscent of the ladder
architecture presented here. The model is otherwise very elegant but
is prone to overfit its lateral connections and exaggerate the
feedback loops between units. By using the same tricks as here,
injecting noise for the benefit of learning lateral and top-down
denoising, it might be possible to learn the lateral connections
reliably.

\section{Conclusions}

In this chapter, a ladder network structure was proposed for
autoencoder networks.  The network's lateral shortcut connections give
each layer the same representational capacity as stochastic latent
variables have in hierarchical latent variable models. This allows the
higher levels of the network to discard information and focus on
representing more abstract invariant features.

In order to support efficient unsupervised learning in deep ladder
networks, a new type of cost function was proposed.  The key aspect is
that each layer of the network contributes its own terms to the cost
function.  This means that every mapping in the network receives
training signals directly from some term which measures local
reconstruction errors.  In addition to the immediate training
information, the network also propagates gradient information
throughout the network.  This means that it is also possible to add
terms which correspond to supervised learning.

The price to pay is that each higher-level cost function needs to be
matched with a decorrelation term which prevents the representation
from collapsing.  This is analogous to the competition used in
unsupervised competitive learning.  Additionally, it is often useful
to decorrelate the inputs because otherwise the network is biased
towards finding a PCA solution.

Preliminary experiments verified that the network was able to learn
abstract invariant features and that the extra terms in the cost
function speed up learning. The experiments support the notion that
the model scales to very deep models and works well together with
supervised learning but much larger experiments are still required to
verify these claims.

\section*{Acknowledgments}

I would like to thank Tapani Raiko, Antti Rasmus and Yoshua Bengio for useful
discussions.  Antti Rasmus has been running experiments in parallel to
this work and his input on how different versions performed has been
invaluable.  Jürgen Schmidhuber, Kyunghyun Cho and Miquel Perelló Nieto
have made available collections of citations which has saved plenty of my
time when preparing the manuscript.

Last but certainly not least, I would like to thank Erkki Oja for
creating the environment where the ideas which underly the work
presented here have been able to develop.  Erkki has always supported
my research. His example has shown how it is possible to follow
intuition in designing unsupervised learning algorithms but he has
also always emphasized the importance of rigorous analysis of the
convergence and other properties of the resulting algorithms.  Without
this combination and his pioneering work in neural PCA, nonlinear PCA
learning rule and ICA, none of the research reported here would have
gotten very far.

\printbibliography

\end{document}